\pdfoutput=1

\documentclass[11pt]{article}

\usepackage{acl}

\usepackage{times}
\usepackage{latexsym}

\usepackage[T1]{fontenc}

\usepackage[utf8]{inputenc}

\usepackage{microtype}

\usepackage{enumitem}

\usepackage{tabularx}
\usepackage{stfloats}
\usepackage{multirow}
\usepackage{graphicx}

\title{Automatic Generation of German Drama Texts \\ Using Fine Tuned GPT-2 Models}

\author{Mariam Bangura, Kristina Barabashova, Anna Karnysheva, Sarah Semczuk, Yifan Wang \\
  Universität des Saarlandes \\
  \texttt{ \{maba00008, krba00001, anka00001, s8sasemc, yiwa00003\}@stud.uni-saarland.de}
  \\
  \texttt{ 7009604, 7023878, 7010958, 2573377, 7023035}}

\date{4/11/2022}

\begin{document}
\maketitle
\begin{abstract}
This study is devoted to the automatic generation of German drama texts. We suggest an approach consisting of two key steps: fine-tuning a GPT-2 model (the outline
model) to generate outlines of scenes based on keywords and fine-tuning a second model (the generation model) to generate scenes from the scene outline. The input for the neural model comprises two datasets: the German Drama Corpus (GerDraCor) and German Text Archive (Deutsches 
Textarchiv or DTA). In order to estimate the effectiveness of the proposed method, our models are compared with baseline GPT-2 models. Our models perform well according to automatic quantitative evaluation, but, conversely, manual qualitative analysis reveals a poor quality of generated texts. This may be due to the quality of the dataset or training inputs.
\end{abstract}

\section{Introduction}
Text generation is a subarea of natural language processing (NLP), appearing in the 1970s \cite{goldman1974computer}. Its main purpose is the automatic generation of natural language texts, which can satisfy particular communicative requirements \cite{liu2009encyclopedia}. Text generation can be a constituent of AI-based tools related to machine translation, dialogue systems, etc. Computational generation of stories is specifically challenging task, as it refers to the problem of selecting a sequence of events or actions that meet a set of criteria and can be told as a story \cite{alhussain2021automatic}. Many studies focus on automatic story generation \cite{cheong2014suspenser}, however, a limited number of them emphasize drama generation \cite{theaitre2020}.

Dramatic texts differ from other genres by having dialogues of acting characters, authorial notes, scenes, and other specific elements, usually written for the purpose of being performed on stage \cite{lethbridge2004basics}. Therefore, the methods described in research devoted to generation of narratives or poetry is not always applicable for the drama generation. The approaches considered in our study are mentioned in Section \ref{relatedwork}.

Nowadays, some of the most advanced methods for text generation comprise transformer decoder or encoder-decoder architecture pre-trained on large-scale unsupervised texts. In previous study referring to drama generation, GPT-2 is applied \cite{theaitre2021}. In the current study, we propose an approach to the generation of drama texts in German, based on the production of outlines \cite{fan-etal-2018-hierarchical, yaoetal}, and compare it with two baseline GPT-2 models. The detailed information about these models and their comparison can be found in Section \ref{approach}. The datasets, used as training materials for the system, are described in Section \ref{preprocessing}.

In order to analyze the performance of story generation models, various evaluation metrics can be involved \cite{alabdulkarim2021automatic}. For the models represented in the current study, we propose automatic quantitative evaluation along with manual qualitative analysis, described in Section \ref{results}. The main challenges and limitation referring to the proposed approach and ideas for further improvement of drama generation are discussed in Section \ref{discussion}.

\section{Related Work}
\label{relatedwork}

Automatic text generation has long been a task of research interests, and various approaches have been proposed to improve the quality of generated outputs. Among all genres, story generation sees the most innovation and progress. Before the era of deep learning, some structural and planning-based models have been applied to perform story generation. The prevalence of RNN \cite{rumelhart1986learning} and LSTM \cite{hochreiter1997long} motivated researches to introduce deep learning to the field of text generation, which results in higher model capacity and better performance. Leveraging language models with more complex architecture and pre-trained on large scale datasets further improved the generation quality by a considerable margin. \cite{alabdulkarim2021automatic}

In addition to the increasing complexity of model architecture, researchers are also committed to proposing innovative generation schemes. \citeauthor{peng2018towards} attempted to steer generation by adding control factors. They extracted control factors from existing corpora and trained a model conditioned on them, so that users can control the generation process by selecting different control factors. \citet{fan-etal-2018-hierarchical, fan2019strategies} explored the possibility of a hierarchical story generation process, where an intermediate stage expands the given prompt and simplifies the following generation process by conditioning it on expanded prompts. Similarly, \citet{consistency-enhanced-story-generation} also applied a two-stage story generation scheme, where the system additional generates a story outline as a guideline for the second stage. It is shown that the hierarchical generation scheme effectively enhances the consistency and coherency of outputs. 

Despite the similarities with story generation, drama generation faces some extra challenges. Firstly, a drama play is usually longer than the upper limit of pre-trained language models, thus an iterative generative process is necessary. Secondly, the lack of prompt-output data makes it impossible to adopt the same approaches as in story generation, and the model must learn to generate plays from nothing. The inherent difficulty of drama generation task discourages researches in this field. To our best knowledge, the only drama generation model is THEaiTRE project \cite{theaitre2020, theaitre2021}. The system leverages a GPT-2 model to generate each scene step by step conditioned on both local and remote contexts. However, the generative model is not fine-tuned on any drama texts, and the generation process requires intensive human interference, which compromise usability of the model and is not suitable for amateur users.

\section{Drama Preprocessing}
\label{preprocessing}

 \subsection{Corpora}
\begin{table*}[hpt]
\small
\centering

    \begin{tabularx}{\textwidth}{l X X}
    \hline
    \multicolumn{1}{l}{\textbf{Method}}
    & \multicolumn{1}{l}{\textbf{Parameters}} 
    & \multicolumn{1}{l}{\textbf{Functionality}}\\
    
    \hline
    
    \_\_init\_\_
    
    & {\textbf{output}}: the empty dictionary that is filled with the data from processed XML file
    & - initializes instant variables used for the XML tags and processed text\\ 
    &
    & - assigns the empty “output” dictionary to the instance variable\\
    
    \hline
    
    startElement
    
    & {\textbf{xml\_tag}}: the start xml-tag (of the <tag> form) which is passed to the method from the file
    & - stores xml-tag and its attributes in instance variables\\
    & {\textbf{attrs}}: attributes of the tag\\

    \hline
    
    endElement
    
    & {\textbf{xml\_tag}}: the end xml-tag (of the </tag> form) which is passed to the method from the file
    & - stores the text processed between start and end tags into a specific instance variable\\
    
    \hline
    
    characters
    
    & {\textbf{content}}: the text between start and end xml-tags
    & - processes the text by skipping empty lines, tokenizing text into words at spaces\\
    
    & 
    & - normalizes words spelling if needed (in GerDraCor only)\\
    
    &
    & - stores processed words by adding them into a list\\

    \hline
    \end{tabularx}

\caption{XMLHandler Class Structure}
\label{tab:xmlhandler}
\end{table*}

The input for the neural model were dramas from the German Drama Corpus (GerDraCor) developed by the Drama Corpora Project (DraCor) \cite{fischer_frank_programmable_2019} and German Text Archive (Deutsches Textarchiv or DTA) \cite{DTA_2022}. 

GerDraCor consists of 591 German dramas, with the earliest written in the 1640s and the latest in the 1940s. 46 dramas appeared to be the same with the ones in DTA and were removed, resulting in 545 dramas used from GerDraCor. In the corpus, speakers, stages and sets\footnote{Stages and sets are texts describing the setting (decorations, position of characters) or commenting on characters’ actions and manner of speech.}, scenes and acts are annotated. There is also metadata available for the whole corpus and containing information about number of speakers and their sex, number of acts and words, etc.

DTA, hosted by the CLARIN service center at the Berlin-Brandenburg Academy of Sciences and Humanities, is the largest single corpus of historical New High German that contains around 1500 cross-genre texts from the early 16th to the early 20th century. 92 drama texts with an orthographic normalization of historical spelling were extracted from the corpus. One of them was excluded, as it was a poem. All historical spellings are adopted true to the original, i.e., they are not implicitly modernized. However, modern or otherwise normalized equivalents of historical writings may be noted with the tags <orig> (historical spelling) and <reg> (modernized/normalized spelling) \cite{DTA_2022}.

The standard GerDraCor format and DTA basic format (DTABf), which were used in this work, follow the P5 guidelines of the Text Encoding Initiative (TEI), which are specified for the annotation of historical printed works in a corpus \cite{DTA_2022, fischer_frank_programmable_2019}. The TEI Guidelines for Electronic Text Encoding and Interchange determines and document markup languages for the representation of the structural and conceptual text features. They refer to a modular, extensible XML schema, consisting of a set of markers (or tags) and accompanied by detailed documentation, and they are published under an open-source license\footnote{https://tei-c.org/}.

The following sections describe how dramas from aforementioned sources were parsed and preprocessed in Python.

\subsection{Drama Parsing}
Parsing of dramas in XML format was performed with XMLHandler class inheriting from ContentHandler class from “xml.sax” module. This class reads xml-tags and operates with their parameters and/or content between starting and closing tags. The class contains methods that were overwritten in order to suit the task of parsing dramas from both GerDraCor and DTA (Table \ref{tab:xmlhandler}).

The tag passed to “startElement” and “endElement” defined how the content between tags should be stored. For example, if “startElement” read <TEI> tag, then the value of the “xml:id” tag was stored from that as drama id; if a tag “</text> was passed to the “endElement”, then it signaled of the end of the drama, and stored all the previously parsed text in a dictionary under the drama id as a key. The text itself was the content read and written in “characters” method and could be the speech of a particular character between specific opening and closing “speech tags”, or, similarly, a description of a stage or a set. Additionally, inside “characters”, text was orthographically normalized: historical spelling of words was replaced with modern spelling, which was looked up in a file containing obsolete-modern spelling pairs and was produced earlier with a File Comparator (described in detail in Section \ref{sec:comparator}). That was done for GerDraCor exclusively, as DTA already contained normalized versions of dramas.
In general, XMLHandler was designed to go through each drama, and extract all the drama text, excluding the front page and the cast list. Further, parsed dramas were consequently written into a single text file. In order to separate dramas and their parts from each other, specific tags were introduced: “\$” as opening tag and “@” as a closing tag, which were followed by the attribute name or value without a blank space. For example, at the start/end of each drama a line with an opening/closing tag and drama id was written (e.g., “\$id\_ ger000569” at the beginning and “@id\_ger000569” at the end) (Table \ref{tab:parsetags}).

\begin{table*}[hpt]
\small
\centering
\resizebox{\linewidth}{!}{
    \begin{tabular}{llll}
    \hline
    \multicolumn{1}{l}{\textbf{Attribute name}}
    & \multicolumn{1}{l}{\textbf{Text following the “\$” or “@” tag}} 
    & \multicolumn{1}{l}{\textbf{Text enclosed between tags}}
    & \multicolumn{1}{l}{\textbf{Example of opening/closing tag}} \\
    
    \hline
    
    Drama id
    
    & id\_dramaid
    & Parsed drama
    & \$id\_ger000569 / @id\_ger000569\\
    
    \hline
    
    Set / stage$a$
    
    & 
    & A set / a stage
    & \$/@\\

    \hline
    
    Scene/act$b$
    
    & scene
    & A scene / an act
    & \$scene / @scene\\
    
    \hline
    
    Speaker id
    
    & sp\_\#speakername
    & A speech of a particular character
    & \$sp\_\#detlev / @sp\_\#detlev\\

    \hline
    \end{tabular}
}
\footnotesize{$^a$ There was no text following “\$” and “@” signs for sets and stage, and the text was enclosed just between those signs.\\ $^b$ 126 dramas in GerDraCor and 15 dramas in DTA did not contain scenes and were separated by acts or equivalent text delimiters, which were marked with a “scene” tag.}\\
\caption{Tags Used in Parsed Dramas with Examples}
\label{tab:parsetags}
\end{table*}

\begin{table*}[hpt]
\small
\centering

    \begin{tabularx}{\linewidth}{l| X| X| X| X| X| X| X| X}
    \hline
    
    {\textbf{transliterated}}	& Hinweg &	sie &	nah’n &	Dort &	sind &	wir &	sicher & \\	
    {\textbf{normalized}} &	Hinweg &	sie	& nah &	‘n &	Dort &	sind &	wir &	sicher  \\

    \hline
    \end{tabularx}
\caption{Example of Erroneously Added Blank Space After Normalization}
\label{tab:blank_space}
\end{table*}

The function for writing parsed drama allowed to produce two different outputs: dramas with the whole text parsed or only characters' speeches (separated by scenes as well) without sets or stages. Eventually, the latter version was used for the further model training. Figure \ref{fig:parsed_ex} shows an example of a drama parsed from GerDraCor with characters’ speeches alone.

\begin{figure}[!htp]

\noindent\fbox{\begin{minipage}{18em}

\$id\_ger000066

…$^a$

\$scene

$^b$

\$sp\_\#dalton

Ein abscheuliches Unglück – ich kann es nicht erzählen – dieser Tag ist der letzte
dieses Hauses.

@sp\_\#dalton
\\

\$sp\_\#frau\_von\_wichmann

Dalton – ist es –

@sp\_\#frau\_von\_wichmann
\\

\$sp\_\#dalton

Belmont –

@sp\_\#dalton
\\

\$sp\_\#frau\_von\_wichmann

Ach – lebt meine arme Julie noch?

@sp\_\#frau\_von\_wichmann

…

@scene

…

@id\_ger000066
\end{minipage}
}

\footnotesize{$^a$ ”…” replaces the text skipped in this example.\\ $^b$ Blank lines are added for the convenience of reading the example.}
\caption{A Shortened Example of a Drama Parsed Only with Speeches}
    \label{fig:parsed_ex}

\end{figure}

\subsection{File Comparator}
\label{sec:comparator}
Since it was undesirable for generated dramas to contain antiquated spellings and characters, the version of DTA texts used for training the model was the normalized version offered by the resource. GerDraCor did not offer normalized versions of their drama texts, though. To mitigate the influence of historical spelling on the training of the model, an effort was made to normalize GerDraCor texts by using DTA texts.

The DTA offers different versions of each of their drama texts, two of which were important for the File Comparator.

\font\vse=cmr10 at 5pt
\setbox0=\hbox{a}
\begin{enumerate}
    \item \textbf{transliterated}: A character-normalized version with transliterated orthography. Given the age of many of the dramas, the original texts included characters outside the Latin-1 encoding, as for example the 'langes s' (U+017F) or the elevated 'e'(U+0364) for marking umlauts.
    \item \textbf{normalized}: A version standardized with regard to spelling, as well as transliterated orthography. Historical spellings such as "Erk\raise\ht0\rlap{\hbox to \wd0{\hfil\vse e\hfil}}\box0ndtnuß." and "weißheyt" are transferred to their modern equivalents "Erkenntnis" and "Weisheit".
\end{enumerate}

Therefore, a collection of word pairs was created, by comparing the transliterated and the normalized versions of the DTA drama texts (Table \ref{tab:transnorm}). Punctuation and other unwanted characters (e.g., “\%”, “(“, “/”) were cleaned from the strings before comparison. Each word pair consists of the old spelling of a word, as well as its modern equivalent. Using this list of word pairs, words in GerDraCor with the old spelling could be changed into their new form.

Since the normalized version resolved hyphenation at the page and line break and sometimes replaced one word with two words, or connected two words into one, the word pairs could not be collected by simply comparing each line word by word in both version. Sometimes, it was indicated in the DTA normalized version, if words were previously merged (e.g., “wie\_es” in the normalized version corresponded to "wie's" in the original text). However, such indication was not done consistently: “thu’s” for example was normalized into "tu es" without an underscore, and therefore, could be treated by the algorithm as two words rather than a single unit.

Issues like these could be easily solved by checking for a specific pattern. The algorithm detects words ending with “‘s” in the transliterated version and tests whether the corresponding word in the transliterated version is followed by an “es”, and if this is the case, then the normalized version likely contains two words (e.g., transliterated “thu’s” is correctly paired with the normalized "tu es").  But sometimes the normalized version added spaces between words, which could not be predicted and caused wrong indexing, meaning two different words in the line to be compared to each other, as shown in the example in Table \ref{tab:blank_space}. Added space in the normalized version (“nah ‘n”) causes the algorithm to combine wrong words in pairs, e.g., “Dort – ‘n”, meaning that “’n” is considered a normalized version of “Dort”.

In order to exclude wrong pairs, where two different words were treated as normalized and transliterated versions of the same word, an algorithm to compare the similarity of words was implemented. If the normalized version was too different from the transliterated version, the word pair was considered faulty (consisting of two different words). 
Firstly,  Levenshtein Distance was used to find possibly faulty word pairs. With using similarity threshold of 3, which appeared to be the most optimal threshold, this method excluded 576 word pairs, but many of them seemed to be correct edits of old spellings (Table \ref{tab:levenshtein}). 

For that reason, it was decided to try another method and estimate word similarity in each pair with the SequenceMatcher class from the “difflib” module. SequenceMatcher uses “Gestalt Pattern Matching” algorithm for string matching. In case, similarity ratio between words in a pair was less than 0.5\footnote{Ratio varies from 0 to 1, where 0 means no commonalities and 1 means identical strings.}, this word pair was deleted from a list of transliterated-normalized pairs. As getting rid of wrong pairs was the priority, the 0.5 threshold allowed us to exclude as many as possible faulty pairs at the cost of losing a few correct ones. Although, this method excluded 712 pairs (more than Levenshtein distance), more of them looked like real faulty pairs (Table \ref{tab:seqmatcher}).

Thus, the final version of File Comparator normalizes words by using word pairs left after excluding faulty word pairs with SequenceMatcher.

While parsing GerDraCor, if the word from drama was found in the dictionary of word pairs, it was lowered, changed to its normalized version and restored with regards to its original capitalization.

\begin{table}[hb]
\small
\centering

    \begin{tabularx}{\linewidth}{X X}
    \hline
    \multicolumn{1}{l}{\textbf{Transliterated}}
    & \multicolumn{1}{l}{\textbf{Normalized}}\\
    
    \hline
    
    Ueberraschungen	& Überraschungen\\
    Medicinerei	& Medizinerei\\
    practicieren &	praktizieren\\
    Caffeegeschirr &	Kaffeegeschirr\\
    Cigarettentasche &	Zigarettentasche\\
    Hausflurthür &	Hausflurtür\\
    Nachtheil &	Nachteil\\
    Legirung &	Legierung\\
    legirt &	legiert\\
    Gratulire &	Gratuliere\\
    nothwendigerweise &	notwendigerweise\\
    adressirt &	adressiert\\
    cuvertiert &	kuvertiert\\
    todtgeboren &	totgeboren\\

    \hline
    \end{tabularx}

\caption{Examples of Pairs Collected from Transliterated and Normalized Versions of DTA Drama Texts}
\label{tab:transnorm}
\end{table}

\begin{table}[ht]
\small
\centering

    \begin{tabularx}{\linewidth}{X X}
    \hline
    \multicolumn{1}{l}{\textbf{Transliterated}}
    & \multicolumn{1}{l}{\textbf{Normalized}}\\
    
    \hline
    
    Wohlhäbige	& Wohlhabende\\
    Verlaubst &	Laubest\\
    Thu’s &	tue es\\
    daß’s &	dass es\\
    hoamgangen &	heimgegangen\\
    Zen &	Zähne\\
    veracht’ &	Acht\\

    \hline
    \end{tabularx}

\caption{Examples of Word Pairs Excluded After Checking for Faulty Word Pairs with the Levenshtein Distance Algorithm}
\label{tab:levenshtein}
\end{table}

\begin{table}[ht]
\small
\centering

    \begin{tabularx}{\linewidth}{X X}
    \hline
    \multicolumn{1}{l}{\textbf{Transliterated}}
    & \multicolumn{1}{l}{\textbf{Normalized}}\\
    
    \hline
    
    Hizt &	Jetzt\\
    nachi &	nage\\
    itz	& Jets\\
    Creyß &	Kreis\\
    Flick &	Flügge\\
    Vehd &	Fett\\
    dy	& die\\

    \hline
    \end{tabularx}

\caption{Examples of Word Pairs Excluded After Checking for Faulty Word Pairs with the SequenceMatcher Algorithm}
\label{tab:seqmatcher}
\end{table}

\section{The Proposed Approach}
\label{approach}

Inspired by the two-stage story generation approaches employed by \cite{fan-etal-2018-hierarchical, yaoetal}, we also decided to divide the drama scene generation process into two stages. However, while \citeauthor{fan-etal-2018-hierarchical} first generate a storyline, which is subsequently used as input to the model that generates the story, we train a model to produce outlines, which become part of the input prompt in the second stage. Likewise, our approach is different from \citeauthor{yaoetal}'s in that it uses just 10 keywords instead of one keyword per sentence in the story. With this approach, we aim to guide the generation process of the model by providing it with the keywords summarizing the most important parts of each scene. Our second goal is to reduce the workload of the user by allowing them to provide only 10 keywords and let the hierarchical model do the rest of the work.

First, we fine-tune a GPT-2 model (the \textit{outline model}) to generate outlines of scenes based on an input of keywords extracted from the text. In the second step, we fine-tune a second model (\textit{the generation} model) to generate scenes based on input which consists of the outline of the scene, a summary of the remote context as well as that of the local context.

\subsection{GPT-2}
GPT-2 \cite{Radford2019Gpt2} has been demonstrated to achieve state-of-the-art results in a range of NLP tasks such as natural language inference, semantic similarity, text classification as well as question answering. Moreover, GPT-2 has successfully been used for story generation \cite{consistency-enhanced-story-generation,seeetal}.

GPT-2, introduced by \citeauthor{Radford2019Gpt2}, is an auto-regressive transformer consisting of 12, 24, 36 or 48 decoder blocks, depending on the size of the model. In contrast to BERT \cite{bert}, which consists of encoder blocks only, GPT-2 stacks decoder blocks. Furthermore, an important property of GPT-2 is its autoregressivity, i.e. the model conditions the next token on the previous token thus allowing text generation.

According to \citeauthor{Radford2019Gpt2}, an additional key feature of GPT-2 is its ability to learn a downstream task in a zero-shot manner, i.e. without any need for parameter tweaking or modifications to the architecture of the model.

GPT-2 was trained with a slightly modified language modeling objective: instead of estimating the conditional distribution $P(output|input)$, GPT-2 estimates $P(output|input,task)$. But, instead of separately modeling this at the architectural level, the task can be prepended to the input sequence.

As there is no official GPT2 model for German, we use the German GPT2 model\footnote{https://huggingface.co/dbmdz/german-gpt2} uploaded to Huggingface. It uses the 12-block setting, resulting in a 117M parameter model.
The model was trained on a 16GB and 2,350,234,427 tokens data set consisting of data from the Wikipedia dump, EU Bookshop corpus, Open Subtitles, CommonCrawl, ParaCrawl and News Crawl. 

\subsection{Pre-processing \& Train/Dev/Test Split}
First, we pre-process the Dracor dataset, generating training instances needed for training both models. As both the outline and the generation models use scenes and gold standard outlines as input, we generate those first. 

For both \textit{scenes} and \textit{outlines} we create two versions: one with speakers left in the text and one without speakers. The first version serves as input to both models, while the latter is only used once during the keyword extraction process. In the first version, each utterance starts with a <speaker name:> and is followed by a newline character, so that the actual utterance is on a separate line. For the version without speakers, we simply make sure each utterance is on a separate line. For sentence boundary detection, we employ the NLTK tokenizer for sentences from the NLTK Tokenizer package \footnote{https://www.nltk.org/api/nltk.tokenize.html}.

In addition, as there are no real outlines available for the plays, we experiment with two summarization algorithms to get the \textit{gold standard outlines}. First, following \citeauthor{consistency-enhanced-story-generation}'s approach we employ TextRank, an extractive text summarization algorithm, to extract the outlines from scenes. We also try abstractive summarization with a BERT2BERT model\footnote{https://huggingface.co/mrm8488/bert2bert\_shared-german-finetuned-summarization} trained on MLSUM, a dataset of 1.5M online news articles. Upon inspection, we found that the BERT2BERT model's output was unsatisfactory: most of the time the summary consisted of 2-3 sentences and was often truncated. Furthermore, as the format of a play presupposes some form of a dialogue, it is quite different from that of a normal text written in prose.  We hypothesize that the strange output of the model is due to it having been trained on news articles. Thus, we proceed with utilizing the TextRank algorithm for outline generation. Prior to performing summarization with TextRank, we remove the speakers, and add them back in to each sentence in the outline.

\subsection{The Outline Generation Model}

As the data set we use does not have gold standard outlines, we decided to follow \citeauthor{consistency-enhanced-story-generation}'s approach, in which they use Textrank (\textit{add citation} to extract the outline of the story (or the scene in our case). We then utilize these outlines as the ground truth output for our model. As input to the \textit{outline model}, we use keywords extracted from the scenes and their outlines. 

\subsubsection{Keyword extraction}
In the search for a keyword extraction algorithm which could yield a good set of keywords for each scene/outline, we have experimented with 6 different algorithms: \textit{Yake} \cite{yake}, \textit{Rake} \cite{rake}, \textit{MultiRake} \footnote{https://pypi.org/project/multi-rake/}, \textit{KeyBert}  \cite{keybert}, \textit{TextRank}\cite{textrank} and \textit{tf-idf}.

\paragraph{Keyword Extraction Algorithms}
\textit{RAKE} first generates a set of candidate keywords for the document subsequently creating a co-occurrence matrix from those. In the next step, for each candidate a score, deﬁned as the sum of its member word scores, is calculated. The word scores are calculated using word frequency (freq(w)), word degree (deg(w)),and (3) ratio of degree to frequency (deg(w)/freq(w)).

\textit{MultiRake} is simply the multilingual version of the \textit{RAKE} algorithm which has some additional parameters such the addition of one's own stopwords or the possibility to vary the length and number of keywords.

\textit{KeyBert} is based on creating BERT embeddings for both the individual tokens in a document as well as the document itself. Then, the cosine similarity of the embedding of each word and the document in which the word appears is calculated. Those words that have the highest cosine similarity with the document embedding are identified as the keywords of the document. 

\textit{TextRank} is a graph-based ranking model which takes into account the co-occurrence of words in a window of \textit{N} words, adding edges between those nodes and then applying applying a ranking algorithm until convergence.  

In contrast to the algorithms mentioned above, \textit{tf-idf} not only quantifies the importance of a term to a specific document in a collection of documents but also off-setts it by the number of occurrences of this term in other documents in the set. This allows to mitigate the effect of highly frequent terms occurring in a large number of documents on the final score.

\begin{equation}
    tf(t,d)= \frac{f(t,d)}{\sum\limits_{t^\prime \in d}{f(t^\prime,d)}}
\end{equation}

\textit{YAKE} differs from the other algorithms in that it relies on a set of features which are supposed to characterize each term. These include \textit{casing}, the \textit{position} of the word in the document, \textit{word frequency}, \textit{word relatedness to context} and \textit{frequency of word in  different sentences}. Finally, these features are combined into a single score which represents the word (\textit{S\textsuperscript{w}}).

\begin{equation}
    S(kw) = \frac{\prod_{w\in{kw}}S(w)}{TF(kw)*(1+\sum_{w\in kw}S(w))}
\end{equation}

\paragraph{Algorithm Comparision}
In order to select the most suitable algorithm for this task, we performed a qualitative evaluation of the keyword extraction results. We used a small set of 5 randomly chosen scenes. Upon inspection of the extracted keywords, we observed that only the keywords obtained using \textit{tf-idf} and \textit{TextRank} actually yielded acceptable results. For example, \textit{Rake}, \textit{MultiRake} and \textit{YAKE} return quite a few repeating keyword or keywords or keywords that differ only in the grammatical case. Furthermore, some keywords were simply a concatenation of neighboring tokens which do not make much sense when put together, especially if the preceding tokens are missing. In addition, \textit{RAKE} and \textit{MultiRAKE} return lowercased version of the keywords, which can be problematic for German text, as casing signals the POS of a word and thus serves an important function, distinguishing nouns from 
other parts of speech. As GPT-2 uses byte-pair-encoding, the starting vocabulary, i.e. the set of all individual characters, consists of both lower and upper case characters. This means that when the BPE algorithm learns to merge adjacent characters, it treats \textit{AB} and \textit{ab} as different tokens.

In light of our observations, we decided to extract keywords using \textit{tf-idf} and \textit{TextRank} and train two outline models. 

\paragraph{Keyword extraction}
As a large number of scenes are quite long and the keyword extraction algorithms often return phrases that are only uttered once by the speaker, we decided to try out keyword extraction from both \textit{whole scenes} and \textit{outlines of scenes}. We have noticed that keywords extracted from outlines are often more relevant to the outline. As a result, our models are trained on \textit{keywords extracted from outlines}, where the outline version is that without speakers.

Another important parameter for our keywords input is the number of keywords (\textit{k}) to be extracted from the scenes. Our experiments have shown that when \textit{k} > 10, many of the terms in the lower half of the keyword list are extremely random and unrelated to the outline. As a result, we chose \textit{k}=10 for both \textit{tf-idf} and \textit{TextRank}.

\paragraph{Tf-idf}
Despite using existing implementations of \textit{tf-idf}\footnote{https://scikit-learn.org/stable/modules/generated/sklearn.
feature\_extraction.text.TfidfVectorizer.html} and \textit{TextRank} \footnote{https://pypi.org/project/pytextrank/}, we had to apply some pre-processing steps.
In the case of \textit{tf-idf}, we first apply a \textit{SpaCy}\footnote{https://spacy.io/} POS tagger with the \textit{de\_core\_news\_sm} German model in order to exclude auxiliary verbs, particles, adpositions and adverbs. In addition, any tokens appearing in NLTK's stopword list for German are dropped.

\paragraph{TextRank}
Similarly, we only keep keywords that are not part of the list of German stopwords. In addition, as \textit{TextRank} extracts sequences of tokens, not individual tokens, repetitions containing tokens that only differ by grammatical case are inevitable. In this case, we discard repeated keywords. For instance, in the case of \textit{die gute Oma} and \textit{der guten Oma} we only keep the lemmatized version of the first occurrence of the keyword.

\subsubsection{Model training}
To fine-tune the German GPT-2 model to produce outlines given keywords as input, we concatenate the keywords \textit{K} and the corresponding outline \textit{O} extracted using TextRank and separate them with the \textit{<SEP>} token. In addition, the concatenated sequence \textit{C} is prepended with a \textit{<BOS>} token and a \textit{<EOS>} token is attached to the end of the concatenated input.

The model is trained for 3 epochs with a training batch size of 4 and a test batch size of 8. The default optimizer AdamW is used and the number of warm up steps for the learning rate scheduler is set to 500. The model is evaluated every 400 steps. During training, we compute the \textit{cross-entropy} of the tokens in \textit{C}. 

At test time, the model is fed the sequence \textit{<BOS>} + \textit{K} + \textit{<SEP>} and is expected to generate the outline tokens. Generation stops once the \textit{<EOS>} token is generated. We use \textit{top p sampling}, wherein the next token to be generated is selected from the vocabulary items that make up 70\% (top\_p=0.9) of the probability mass. In addition, \textit{repetition\_penalty} is set to 2.0.

As has been mentioned before, we have trained two versions of the outline model (using the same settings): one in which the keywords are extracted using \textit{tf-idf} and the other using \textit{TextRank}. The two models are evaluated with respect to their performance on the downstream task of scene generation, discussed in Section \ref{sec:quanresults}.

\subsection{The Generation Model}
In the second part of our system, another fine-tuned GPT-2 model is leveraged to generate a drama scene from given start and outline. The generation model can be characterized by the following three aspects:
\begin{enumerate}
   \item Iterative generation: As many drama scenes are longer than the upper length limit of GPT model (1024 tokens), it is not possible to generate a whole scene at once. Therefore, we adopt an iterative generation strategy: in each iteration, the model only generates 100 tokens, and all generated tokens are then fed into a summarizer to produce the prompt for next iteration.  
   \item Dynamic prompt: In our system, the prompt is split into three individual parts: outline, summary of remote context and local context. The outline is either drawn from the original play or generated by the first part of the system, and remains unchanged in all generative iteration. When the outline and generated outputs are longer than 924 tokens, only the nearest 250 tokens are preserved, and the remote context is summarized by a TextRank model. The three parts are concatenated with a <SEP> token to form the prompt of each generation step. In this way, our model can maintain local coherency as well as memorizing important information even if it is mentioned far ahead of the current position. The introduction of outlines provides a guideline for the plot and guarantees global consistency. Thus, the model is provided with dynamic prompt with different information in each iteration. A figure describing the model structure can be found below. (Figure \ref{fig:model_architecture})
   \begin{figure}[htp]
    \centering
    \includegraphics[width=7.5cm]{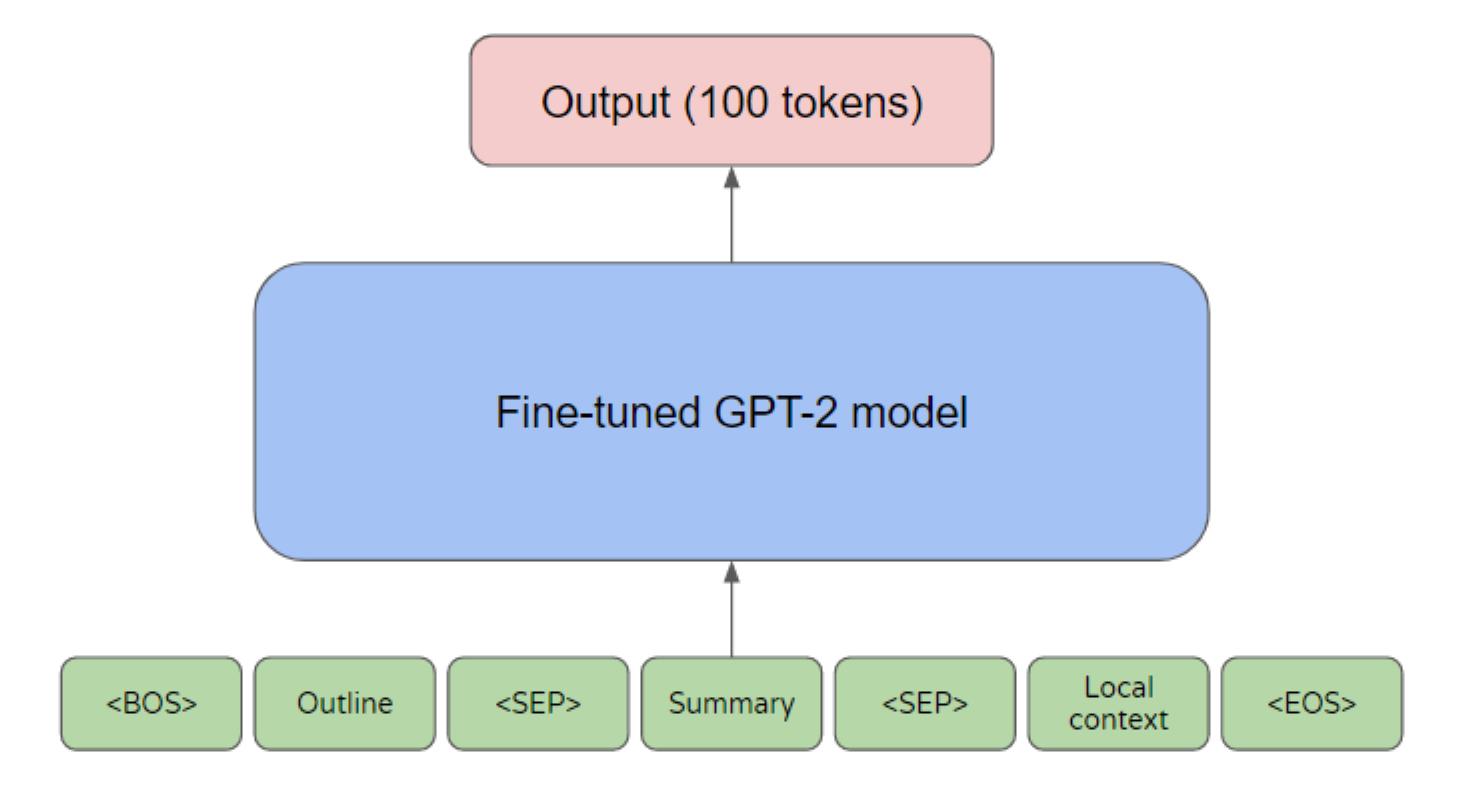}
    \caption{Model architecture }
    \label{fig:positional_bias}
\end{figure}

   \item Automatic post-editing: Despite the improved performance of our fine-tuned GPT-2 model, it still fails to produce drama as human experts. This can be attributed to the inherent difficulty of drama generation and the diverse writing styles and formats of our collected training corpora. To address some recurring format problems, we apply a few automatic post-editing methods. In particular, we have resolved the following issues:
   \begin{itemize}[leftmargin=0pt]
       \item Repetitiveness: As the input information is relatively little at the beginning of generation, the model tends to repeat sentences from prompt or generated lines. To counter repetitiveness, we set the repetition penalty to 1.01 and forbid repeated 4-grams during generation. We also discard any new lines (excluding character names) that have already been generated. Since we have a strict penalty for repetition, it can occur that the model cannot generate a valid line in an iteration. To prevent these cases, the model returns 10 sequences each time for the post-editing module to select, and it is backed off to generation without outline when none of the 10 returns include any valid lines.
       \item Bad character names: In most cases the model is able to identify characters in the play and continues the dialogue with their names. However, it sometimes misspells names or adds new characters abruptly, which harms the plot consistency. In our system we identify misspelling by its edit distance from any given character name. If the edit distance is small (less than or equal to 2), it is considered as misspelling and the wrong name is corrected to a given character name. Otherwise, the new name is seen as an invalid character and will be removed along with its speech.
       \item Empty speeches: The model may output character names at the start of a new line but does not assign any speech to them. We manage to resolve this problem by identifying character names followed immediately by another name and discarding the lines.
   \end{itemize}
\end{enumerate} 

\section{Experiments and Results}
\label{results}
To study the effectiveness of our proposed approach, we compare our models with baseline GPT-2 models. In particular, we have two baseline models: a not fine-tuned GPT-2 model and a GPT-2 model fine-tuned on the same training set but with no outline or summarization (-dynamic prompt). All models are based on a German-language GPT-2 model named german-gpt2 from HuggingFace.\footnote{https://huggingface.co/dbmdz/german-gpt2}

For generation model training, we first extract prompt-generated output pairs from collected corpora and fine-tune our model on them. In particular, the outline part in the prompt is extracted from the original scene using TextRank. We run the training for 3 epochs with a batch size of 8, evaluating on the dev set every 400 steps. A default optimizer is used with 500 warm-up steps and the checkpoint with the lowest perplexity on dev set is chosen, with an early stopping patience of 10. The fine-tuned baseline model is trained on the raw drama scripts directly with the same set of hyperparameters as generation model training, except that it is trained for 10 epochs, as there are fewer optimization steps in each epoch compared to generation model fine-tuning.

\subsection{Evaluation Metrics}

To evaluate the performance of our approach, we adopt several automatic quantitative evaluation metrics as well as a manual qualitative analysis. 100 scenes from test set are generated by each model given start of the scene (approximately 100 tokens) as well as an outline (approximately 200 tokens, set to NULL for baseline models) and their performance is measured and compared by the following metrics.

\begin{itemize}
    \item \textbf{Average number of sentences per speech}: In general, drama is comprised of conversations, which means each character is supposed to take turns to give their speeches. Thus, it is important that the model should not generate a text where only one or two characters give very long speeches. Average number of sentences per speech is a metric reflecting how well the generated plays resemble a human written play in format. Abnormally high value in this metric indicate that model fails to capture the format features of drama.
    \item \textbf{Average sentence length}: Average sentence length is a simple yet effective measurement of performance of generative models\cite{kincaid1975derivation, roemmele2017evaluating}. While too long sentences might harm readability, too short sentences are more likely to be incorrect or illogical in the context. In our experiment, we compare the average sentence length of generated texts to that of human written scripts, to evaluate and compare how each model performs in generating fluent and readable sentences.
    \item \textbf{Perplexity}: We also measure the perplexity score of the generated scenes from each model (including human written plays) using german-gpt2. Perplexity is usually assigned by a to-be-evaluated language model on a real text \cite{jelinek1977perplexity}, while in our case we reverse the process and leverage a pre-trained language model to evaluate the fluency and coherency of generated texts. In particular, to balance the evaluation efficiency and accuracy, we use a stride of 100. Lower perplexity score indicates better coherency.
    \item \textbf{N-gram overlap}: For n=1,2,3, we measure the F1 score of n-gram overlap between the start of scene and generated text. Low value means lower similarity between the generated text and start of the scene.
    \item \textbf{Topic drift}: In addition to n-gram overlap, we also calculate the overlap for the first half and second half of generated texts separately, and measure the proportion of decrease in F1 as a metric for topic drift. We assume that, if a story is globally consistent, the topic drift should be relatively small, while a story lacking plot consistency tends to have larger topic drift.
    \item \textbf{Distinct-n}: To examine the model’s ability of producing diverse words and phrases, we also compute the average proportion of unique n-gram in the whole generated text \cite{li2015diversity}. Higher proportion of unique n-grams reflects that the model is highly capable of generating unseen words or phrases, either by rephrasing existing expressions or introducing new contents.

\end{itemize}

\begin{table*}[htp]
\small
\centering
\resizebox{\linewidth}{!}{
    \begin{tabular}{lcccccc}
    \hline
    \multicolumn{1}{l}{\textbf{Models}}
    & \multicolumn{1}{c}{\textbf{w/o fine-tuning}} 
    & \multicolumn{1}{c}{\textbf{w/o outline}} 
    & \multicolumn{1}{c}{\textbf{w extracted outline}} 
    & \multicolumn{1}{c}{\textbf{w TextRank outline}}
    & \multicolumn{1}{c}{\textbf{w TF-IDF outline}}
    & \multicolumn{1}{c}{\textbf{Human}}\\
    
    \hline
    Sentences per speech
    
    & 64.74
    & 5.40
    & 4.47
    & 5.97
    & 6.04
    & \textbf{3.21}\\

    Sentence length 
    
    & 4.60
    & 6.89
    & 6.16
    & 6.82
    & 6.60
    & \textbf{8.82}\\

    Perplexity
    
    & 20.58
    & 18.90
    & 19.18 
    & 18.82
    & 19.46
    & \textbf{17.13}\\
    
    \hline 
    1-gram overlap
    
    & 0.11
    & 0.19
    & \textbf{0.20}
    & 0.17
    & 0.18
    & 0.10\\
    
    2-gram overlap
    
    & 0.012 
    & 0.028
    & \textbf{0.040}
    & 0.020
    & 0.022
    & 0.009\\
    
    3-gram overlap 
    
    & 0.0007
    & 0.0053
    & \textbf{0.0109}
    & 0.0024
    & 0.0032
    & 0.0013\\
    \hline 
    Topic drift (2-gram)
    
    & -8.33\%
    & 34.3\%
    & \textbf{18.2\%}
    & 27.3\%
    & 42.3\%
    & 20.0\%\\
     
    Topic drift (3-gram)
    
    & 28.6\%
    & 66.3\%
    & \textbf{23.0\%}
    & 34.6\%
    & 79.2\%
    & 57.1\% \\
    
    \hline
    Distinct-1
    
    & 0.503
    & 0.433
    & 0.463
    & 0.438
    & 0.443
    & \textbf{0.576}\\

    Distinct-2
    
    & 0.880
    & 0.842
    & 0.860
    & 0.846
    & 0.846
    & \textbf{0.921} \\
    
    Distinct-3
    
    & 0.969
    & 0.963
    & 0.966
    & 0.962
    & 0.960
    & \textbf{0.982}\\
    \hline
    \end{tabular}
}
\caption{Automatic evaluation results on 100 test set.}
\label{tab:results}
\end{table*}

\subsection{Quantitative Results}
\label{sec:quanresults}
Table \ref{tab:results} shows the results of our automatic evaluation. It is obvious that the model without fine-tuning fails to produce texts that are formally similar to drama: each speech consists of on average 64.74 sentences and each sentence is composed of only 4.6 words, indicating it just start a speech randomly and many sentences are only one phrase or even one word. For this reason, it will not be analyzed later in this section. All other models perform reasonably well in average number of sentences per speech. The human-written scripts have the lowest value.

Similar patterns can be observed in average sentence length and perplexity. Human-written scripts demonstrate the best readability and coherency. Among the machine generation approaches, despite the gap being trivial, the model with no outline and the model with outline generated by keywords (TextRank) display superiority to the model with extracted outlines in terms of fluency. 

When it comes to n-gram overlap, the model with extracted outlines has by far the highest overlap with the given start of the scene, followed by the model with no outlines. The models with generated outlines do not reach a decent result, probably because of the poorer quality of outlines. It is worth mentioning that the real drama scripts have the lowest overlap score. We attribute this to the outstanding ability of human experts of rephrasing and controlling the flow of plot, thus it is not directly comparable to machine generation approaches.

Besides, we notice that introducing extra outline information indeed contributes to a better global consistency: models using both extracted outlines and outlines generated by TextRank keywords show competitive or even better performance than the human-written plays in topic drift, while model that do not leverage such information suffer severely from topic drift.

Finally, no significant difference is observed in the ability of using diverse vocabulary among machine generation models. Human playwrights, as we have expected, show their irreplaceable advantage in diction.

\subsection{Qualitative Analysis}

\subsubsection{Qualitative Evaluation of Outlines}
Firstly, in most of the cases, only the first line of the outline contains a speaker. Naturally, this makes it impossible for the subsequent generation model not to come up with random characters that do not appear in the outline. Furthermore, after the first couple of sentences, the generated outline 
quite often consists of direct speech followed by a reporting clause (i.e. "sagte der Mann" - "a man said", "fragte er" - "he said"), as can be seen in both generated outlines in Table \ref{table:gen_outl_append} in the Appendix. This is quite surprising, considering that the gold standard outlines do not contain any such text, as all of the drama pieces are in dialogue format. A possible explanation for this could be that the amount of drama texts used for training is insignificant compared to the large amounts of news data the model was pre-trained on. 

\subsubsection{Qualitative Evaluation of drama texts}
Manual evaluation reveals that none of the models were able to produce coherent and meaningful texts. On average the texts created by the model with no outline are shorter compared to the texts from other models, which mostly end after a maximum iteration number is reached. Though all of the models produced texts that ended with the repetition of mildly changed words or phrases, the model using an extracted outline did so more frequently. This can be seen quite well in the generated example text found in Table \ref{table:generated_ex_1_1} Part 3/3 given in the appendix and might explain the extremely low topic drift values for this model.
The two models using generated outlines did not introduce as many new characters and did not switch between speakers as often as the other two models, creating mostly long monologues instead of dialogues.
All of the models overused '>>' and '<<' in normal dialog and started a lot of sentences with a hyphen. Since this is a problem occurring in all of the models, it can be assumed that the varying formalization across different dramas used in the training process caused this issue.
In multiple drama texts two or more following hyphen were used to mark pauses in speech. One example of an excessive use of hyphen in the original drama texts can be found in the excerpt from 'Die Pietisterey im Fischbein-Rocke' given in Figure \ref{fig:excerpt} in the appendix. The models tend to overuse hyphens in a way, that hinders meaningful text generation instead.

\section{Discussion and Outlook}
\label{discussion}
Our proposed method described above is able to handle some known issues like lack of global information. However, drama generation/completion is a challenging task even for human experts, and in our work, there are still some problems remaining unresolved:
\begin{enumerate}
    \item Abrupt ending: Although a special token <endoftext> is added to the end of each scene and used for training, we notice that in most cases the generation only stops when a maximum iteration number is reached. This will lead to an abrupt ending problem. A better method should be explored to provide more control over the story ending without dramatically harming the conciseness of drama.
    \item Non-uniform format: Despite an extra post-editing process during generation, some inconsistency in format is still not completely avoided. Some bad names are not detected as well and are thus kept in the text and compromise readability.
    \item Instability: While some previous works \cite{theaitre2020, theaitre2021} rely on manual intervention to detect and prevent unsatisfactory generation results, we decided to adopt a more convenient fully automatic approach, which inevitably induces accumulated errors and results in instability in generation.
    \item Incompetence of generating a whole play: The proposed model can only generate one scene at a time and cannot produce a whole play. Future work can focus on this more challenging task, for example by introducing an additional layer to the hierarchy, aiming to generate outlines for each scene based on a outline of the whole play and summary of previous scenes.
\end{enumerate}

\section{Conclusion}
\label{conclusion}
This paper compares the quantitative results of different models attempting the generation of German drama texts. Furthermore it explores the opportunity of generating German drama texts with extracted outlines. While the quantitative results of the models suggested sensible outcomes, qualitative analysis of the generated texts found them to be lacking in regards of coherency, meaning and form. A lot of issues can be hypothesized to stem from the varying formalization in the drama texts used in the training of the models and the poor quality of the generated outlines. A bigger and cleaner dataset of German drama texts would be desirable for further testing of generational models.

\bibliography{anthology,custom}
\bibliographystyle{acl_natbib}

\appendix
\clearpage
\section{Appendix}
\label{sec:appendix}
\setcounter{table}{0}
\renewcommand{\thetable}{\Alph{section}\arabic{table}}
\setcounter{figure}{0}
\renewcommand{\thefigure}{\Alph{section}\arabic{figure}}

\begin{table}[hp]
\centering
\begin{tabular}{|c|c| >{\centering\arraybackslash}m{12cm}| } 
 \hline
 \textbf{id} & \textbf{model} &\textbf{Generated outline} \\ 
 \hline
 id\_ger000018\_5 & textrank & Liddy:
Ich muß sagen – ich bin entzückt..?« »Do is doch was zu do«, sagte er zum Käthchel und wies auf die große Verantwortung hin; Herr Littmann meinte so schön es im Scherz ang'scheckt wie eine Maus in einer Mädl'n oder Frauenzimmerin des Herrn von der Hellenen-Pensionärsfeuerwehre mit dem Namen Ihnen 'nen Kauz auszulachen«. „Mein Vater war ein alter Knabe aus dem Dorf Der hat sich für einen Berufsstart als Tischler entschieden Aber da kommt man ja nicht mehr drauf zurück...“ Nun das ist wahrhaftig kein Wortschatzfehler sondern daß mir auch noch ganz ungebildete Leute zuhören können Als ob mich mein Sohn nur um seine gute Laune gebeten hätte wenn meine Tochter einmal über ihm wäre...\\
 \hline
id\_ger000018\_5 & tf-idf & Valeria:
Ich wollte dich nicht verlassen. – Ich habe dir es noch gesagt; ich bin dein Freund und deine Dienerin gewesen!« »Und du hast mich verraten?«, sagte er zu mir mit dem Augenschmalze auf den Lippen in der Sprache seiner eigenen Muttersprache so leise als möglich.« Nun komme endlich nach Hause zurück von diesem Abend hier im Garten meiner Schwester Liddy oder dieser Ihen Tochter Molly... wenns kein Irrtum ist gegen das Schicksal meines Vaters für die gute Sache aufzuwarten?' laß ihn allein sprechen!». 'tschuldigung Vater, aber da war ein Brief an Sie bei Ihrer Ankunft am Bahnhof : Er kam aus Wien ohne Ihre Erlaubnis herüber ins Haus Ihres Sohnes Marquis de La Salle … Und nun kommen sie alle wieder hierher um sich ihren Spaß daran gönnen ; denn wie gewöhnlich haben wir beide uns schon einmal unterhalten über einige Dinge zwischen Ihnen beiden.... man muß doch sagen daß diese Begegnung eine gewisse Wahr heit enthält!!’ Ach was soll nur sein – lassen Sie mal hören!'.Das wird ja wohl niemand sehen wollen ob Ihr Sohn wirklich einen Fehler begangen hat unter seinem besten Namen.Herr Baronin, erlauben Herr Graf Renz-Honau, bitte recht freundlichst unsere Hand deines Mannes küssend anzubezeigen....binde Fräulein Gräfin Marie wird alles verzeihen müssen meine Frage wäre jetzt keine Antwort nötig öffne deinen Mund weit genug.'' / So sagt mein Bruder zum Doktor – das tut weh!', fragt er scherzhaft nach einer Bemerkung des Herrn Kommissars... vielleicht sei der Mann zu jung dafür geworden, weil er nicht anders kann tun als seine Frau lieben Zu machen ihm seinen Kummer verständlich darzustellen als auch seine Leidenschaften zu zeigen.\\
\hline
\end{tabular}
\begin{minipage}[c]{\textwidth}
\hfill
\caption{Examples of outlines generated by the two outline generation models (tf-idf and textrank)}
\label{table:gen_outl_append}
\end{minipage}
\end{table}

\begin{table*}[hpb]
\small
\centering
\begin{tabularx}{\linewidth}{|l|X|} 
 \hline
 \multicolumn{2}{|c|}{id\_ger000275\_37.txt} \\
 \hline
 START\_OF\_SCENE & Rosamunde:
Ja er ist es!
Gelobt sei Gott!\newline
Flodoardo:
Eure Durchlaucht...\newline
Doge:
Seid willkommen, Edler Ritter, tausendmal... Habt Ihr Euer Wort gelöst?\newline
Flodoardo:
Alles, wie Ihr mir geboten, Ist erfüllt; die Republik Frei von jeglicher Gefahr; Und der Himmel lächelt heiter.\newline
Doge:
Und der Mörder unserer Freunde?\newline
Flodoardo:
\\
 \hline
 w/o outline & Flodoardo:
Wir sind gerettet.\newline
Doge:
Der Tag wird kommen, Daß er sich uns naht.\newline
Flodoardo:
Stimme:
Euer Tod ist unser Tod!\newline
Doge:
Gelobt ist Gott!
...\newline
Flodoardo:
Das Glück ruft nach uns!
– Der Ritter Des Todes, Der die Stadt umringt, Der Ritter von Florenz, Der den Kaiser stürzen will!
– Hört mich an!\newline
Flodoardo:
O Gott!\newline
Doge:
Und er, der den Kaiser stürzen wollen –?\newline
Flodoardo:
Er soll sich erheben, Und der Herr ist frei!
Weh ihm, daß er es erfährt!
Ich bitte Euch, laßt ihn, er soll nicht mehr Euch, Ihm allein, sprechen.
Fluch und Haß und Tod sind für uns Leben.
\\
 \hline
 w extracted outline & Flodoardo: Auch dieser Mensch, der mit uns ins Gefängnis kam, Der auf diesen Felsen gesprungen ist, Und sich vor mir entblößte, so als der Retter der Republik?\newline
Flodoardo:
Er war der letzte, der hier saß, bis zum letztenmal.
Ich bin, wer er?
Dogt mir.\newline
Flodoardo:\newline
Doge:
Wer war der erste?
O wie Ihr ihn nennt, den ersten.\newline
Flodoardo:
Wen nannte der König?\newline
Doge:
Mein Herzog.
Dogt nicht das Gerücht.\newline
Doge:
Die Republik hat ihn ausgerufen.\newline
Flodoardo:
Ein Fürst.\newline
Doge: Ein Mann.
– Ihr, mein edler Herr Abbate; Ihr, hochfürstlicher Herr Follo liegt bereits in den Eisen.\newline
Flodoardo:
Was wird aus Eurer Freiheit?
– Contarini, Euer Oberst Frosso liegt schon In den Eisen; Es wird von der Nacht so viel geredet, Die ihn verderben und aufhetzen.\newline
Parozzi:
seht, daß Ihr, das schwöre ich, Den Prinzen Valerio vor euch schützt.\newline
Flodoardo:
Wünschst du, daß der Herzog nicht?
–\newline
Doge:\newline
2-3:
Wünschest du, daß ich es nicht könne?
– Alle anderen?
– Die ganze Republik?
– – Wir?\newline
Flodoardo:
Denselben, den ihr im Dunklen habt?\newline
Doge:
So wie sie sind, so muß ich auch das sagen.
Wir müssen ihn selbst holen.\newline
Flodoardo:
Und der Richter von Florenz?\newline
Doge:
Der Herzog kommt mir zu nah.\newline
Flodoardo:
Ist die Flucht so ein Glück?\newline
Flodoardo:
der Herzog von Parozzi.\newline
Doge:
Er wird mich mit seinen Kindern einsperren.
\\
\hline
\end{tabularx}
\caption{Example texts generated by the different models.}
Part 1/3. Continued on the next page.
\label{table:generated_ex_1_1}
\end{table*}

\begin{table*}[hpb]
\small
\centering
\begin{tabularx}{\linewidth}{|l|X|} 
 \hline
 \multicolumn{2}{|c|}{id\_ger000275\_37.txtr} \\
  \hline
 w TextRank outline & Flodoardo: Sie sind so frei und liebenswürdig?\newline
Flodoardo:
Und unsere ganze Menschheit?
Und die Menschheit...\newline
Flodoardo:
Und die?
...\newline
Doge:
Und was noch wichtiger ist!
Denn in der ganzen Welt, die auch noch so verwegen, und die so viel Geld bringt, die ganze Menschheit noch so viel Geld macht!\newline
Flodoardo:
Die Menschheit ist ein ganz Anderer!
Ihr macht es doch noch schlimmer, wenn Ihr mir die Hand reichen wollt.\newline
Doge:
Wenn Ihr Euch damit nicht zufrieden macht, so muß ich Euch doch den Kopf zerbrechen lassen.
Flodoardo:
Ihr solltet Euch denn nicht mehr ärgern, wenn Euch Euer Leben nicht mehr wert ist.\newline
Doge:
So könnt Ihr Eure Seele zum Tode schlagen.\newline
Flodoardo:
Wie!\newline
Doge:
Sie sind so frei...\newline
Flodoardo:
So, meine Herren, mit Euch muß man nicht mehr um das Leben tanzen, daß man doch gleich zu Grunde kommen will.\newline
Doge:
Und Euer einziges Recht – so!\newline
Flodoardo:
Das Recht –\newline
Doge:
Denkt Ihr?\newline
Flodoardo:
Ja wenn Ihr mich nicht wollt.
Wir sind doch beide nicht frei.
Ihr werdet noch alle Eurer Liebe, Euren Haß, Euren Eifer, Euren Schmerz verlieren.
Ihr seid mein Freund!\newline
Doge:
Und Ihr wollt leben, wo Ihr es wagt.\newline
Flodoardo:
Es gibt Euch doch wohl in der Welt nichts was mehr ein Mann wagen würde.\newline
Doge:
und Ihr seid auch ein Weib!\newline
Flodoardo:
Ich!
– Ihr seid noch nicht frei?\newline
Doge:
Ihr könnt auch für Euren Mann leben.\newline
Doge:
Wenn ich mir so in Eure Hände lerne... wenn Ihr mir nun nicht mehr traut.\newline
Flodoardo:\newline
Flodoardo: ich schwöre Euch auf Euer Recht.
Ihr seid es, Ihr habt es, so will ich es!
– Auch Euch ist es nicht anders möglich?\newline
Flodoardo:
Nein, Euer Herz ist zu schwach.
Es war ja der einzige Grund, Euch hierher zu schicken.\newline
Doge:
ja was hast du da zu tun?\newline
Flodoardo:
Wie?\newline
Doge:
Das ist ganz der Mann in dieser Welt, der ich sein will.
\\
 \hline
 w TF-IDF outline & Flodoardo: Er hat den Mann, der ihn umgebracht hat, umgebracht...
– O, die Herrschaft hat mehr als uns geboten!
Wenn er uns nicht den Mann genommen hat, den er umgebracht hat, und den Leichnam hat er sich selbst um den Hals geworfen.
O nein, man wird uns doch nicht einmal mit dieser Schmach zu helfen wagen.\newline
Flodoardo:
Der Mord ist doch nicht eine Sache, die wir vor der Welt tun könnten... man müßte es doch wohl auch tun!
– und daß er sich hier umgebracht hat!
– daß wir uns mit ihm verbanden?
– daß die Unschuld uns nicht mehr gebiet.
– daß er sich auch in dem Mord einen Mord an einem anderen schuldig gemacht hat?
– er hat uns nicht mit einem Mord gebieten, den der uns nicht mehr wollte.
– daß der Mörder uns erbarmen hat, ist ein Undank.
– weil er keinen mehr in seinem Leben hat?
– daß er eine Mörderbande an den Mörder hat!\newline
Flodoardo:
Das muß ich sagen...
So soll er sich doch nicht schuldig machen...
– und daß alle in dem Leben eines anderen sind!
– warum sollten sie dem anderen noch was zutrauen!
–
– daß sie uns doch nur noch eine Freude machen könnten!
– er ist doch gar nicht einmal auf dem Wege!
– aber er hat doch etwas, was wir uns zutrauen haben!
– oder daß er sich von einer Stunde an selber schuldig macht...
– daß keiner den Mörder hat, der sich ihm den Mord hat nehmen lassen, und der dem Mörder noch ein Mensch ist!
– Und dann wird er sich in die Arme von diesem Mörder gelegt haben.
– dass wir nichts zu tun hätten!
– ob wir denn von unserem Mord wüssten?\newline
Doge:
Und daß diese Mörder von ihrem eigenen Blut sein werden?
– und daß wir es wirklich nicht mehr sein können, dass er uns die Schuld gibt!
– weil uns das Gesetz sagt.
– und daß ich den Mord nicht will!
– – und daß wir uns das nicht selber antun!
– daß er uns nicht will!
?
– dass er den Mord nicht kann, wenn wir ihn nicht kriegen!
– dass sein Blut nicht weiß!
– was er uns tut!
– denn wenn er das nicht kann!
– uns beide!
– in den Tod!
– wenn er sich selbst umgebracht hat!
– nicht mit dem Blut der Kinder?
– wenn sein Herz nicht in dem Blute ist!
– und wir!
\\
 \hline
 \end{tabularx}
 Continuation of Table \ref{table:generated_ex_1_1}. Example texts generated by the different models.
\\Part 2/3. Continued on the next page.
\end{table*}
\hfill

\begin{table*}[hpb]
\small
\centering
\begin{tabularx}{\linewidth}{|l|X|} 
 \hline
 \multicolumn{2}{|c|}{id\_ger000200\_4.txtr} \\
 \hline
 START\_OF\_SCENE & Kauz:
Schön gute Morgen, Frau Nièce!\newline
Frau Von Erbsenstein:
Der Morgen kann gut und schön sein, ich bin aber bös und wild!\newline
Kauz:
Bös, das kann sein, aber wild –?
Im Gegenteil, ich finde, daß dieser Anzug –\newline
Frau Von Erbsenstein:
\\
 \hline
 w/o outline & Frau Von Erbsenstein:
Er ist noch einmal ein Anderer – er ist ein besserer Herr als du.\newline
Kauz: Ist er aber nicht!
Ich bin nicht in der Stimmung, daß ich mir einbilde.
– Die Nacht ist schön, da kommt es mir vor, als sollte ich schlafen.\newline
Kauz:
Du, Marie, wie ist das möglich?\newline
Kauz:
Ich liebe Sie, aber nicht so sehr, als wenn ich mit Ihnen schlafen wollte.
Der Herr von Boskowitz ist ein Mann, was weiß ich!
Und ich soll mit Ihnen schlafen?\newline
Kauz:
Ich werde heiraten.\newline
Kauz:
Das kann ich nicht sagen.
Ich habe noch nie gesagt, daß ich so jung heiraten will.\newline
Marie –
Ich auch nicht, aber ich sage es ganz offen: ich will nicht verheiratet sein.
\\
 \hline
 w extracted outline & Frau Von Erbsenstein:
Es is wieder Zeit, das Gute zu sagen, das zu sagen, was zu sagen – wie ich sagen will.\newline
Kauz:\newline
Frau Von Erbsensteins Sohn:
Ja, richtig, das weiß man im übrigen schon, den Onkel, der hat auch den Titel schon bekommen.\newline
Herr von Erbsenstein.\newline
Kauz:
Keizer.\newline
Kauz:
»Wild«:
Ach!
Frau von Erbsensteins Tochter!
Keizers Tochter?\newline
Kauz:
Nun, dann wäre die Zeit gekommen, daß ich auf die andere Seite hinübergehe, und michFrau Von Rätzer, Frau Von Erbsen!
...\newline
Kauz:
Ich bin ja schon so früh dran, die Nacht werde ich nicht vergessen, wenn ich mich schon auf die Seite setze.\newline
Kauz:
Das geht nicht, Frau Von Rätzern...\newline
Kauz:
Ja so.
Keizer!\newline
FrauVon Erbsenstein!
Kiezer!
Kiezer?\newline
Kauz:
Da.
Kriezer!
Da sind ja die anderen.
Frau Und da sind es die anderen.\newline
Kauz:
Kaizer!
Was geht uns das alles an!
\\
 \hline
 w Textrank outline & Frau Von Erbsenstein:
Ich glaube sie hat ihn ganz richtig, Frau Von Erbsensteins –\newline
Kauz:
Ja so, ja so.\newline
Kauz:
Die ist ja ganz schwarz und weiß wie die Haare von Hunden!
Aber ich finde die Haare doch noch so dunkel!
Ich muß mal schaun!
Brauchst dich nicht zu drehen, Frau Von Zedlitz, das hat nix mit mir zu tun.
Der Herr hat gesagt es wäre ja ein kleiner Mensch von dem ich glaube ich muß es doch nicht sein?
Der Herr hat denn auch gesagt das ist nichts als ein ungebärdriges Ding, das kann doch ein nicht wahr!
Das heißt, du meinst Herr von Zedlitz?
Ja so – was denn?
Das is ein wenig zu viel der Anstand und wie viel Herr von Zadl ist, aber das is eine andere Geschichte.
Aber ich habe ja immer gesagt das is was für die Leute –\newline
Kauz:
die haben gesagt das is nix von Mensch!
– Das muß doch net anders sein als mit mir –
Was is denn?
Bös!
Der Herr kommt schon zurück.
Ich hab es ja schon immer gesagt das war was er!
Ich habe das so satt mit Herrn von Zedl.
Ja das is halt mein Herr!
So!
Du gehst doch zum Teufel, ob ich mich net bewerbe!
Da kann ich nicht widersprechen!\newline
Kauz:
Also du?
Na wenn du meinst nicht?
Du gehst ja!
der is nur 'n bißchen auf deine Finger gekriegt!
Bösl!
Der Bursche is ja schon wieder!
Wenn du dir nachher den Teufel zum Maul hält, wie die anderen.
Ich glaube du solltest der Kerl sein den ich mir ausgedacht habe!\newline
Kauz:
Das is wirklich gut!
\\
 \hline
 w TF-IDF outline & Frau Von Erbsenstein:
Ich muß mit Ihnen ehrlich und herzlich über unsere Lage sprechen, wir habens immer vermieden, daß es zu einer Entscheidung kommt.
Jetzt weiß ich schon, daß der Alte nicht ein ehrlicher Mensch ist, sondern ein Schuft!
Was ist die Frau?
Ich bin auch da!
Frau von Erbsenstein, Sie müssen kommen!
»Ein Herr hat mir erzählt, daß Sie in Amerika sind, wie Sie so schön heißen?
»Ein junger Herr aus der Stadt ist hier von der Börse gegangen.
Und ein so ein alter Mann wie Sie – wie?
...«
Im Gegenteil sind wir doch alle drei ein hübsches Paar!
Frau...
Im Gegenteil!\newline
Kauz:
Und die Herrn!
Sind drei Jahre lang im Ausland – das freut mich!
Frau Soehne!
Frau!
!
Was sagen Sie?
Haben Sie denn das ganze Jahr wohl noch nicht bemerkt?
Es ist doch was anderes!
Ein Mensch!
»Was steht drin?«
Haben Sie das ganze Jahr noch nicht bemerkt.
»Was wissen Sie noch von dem Geld?
Ich hätte es Ihnen sagen müssen, daß ich den alten Herrn aus Amerika bin, den ich noch nicht kenne.
»Und nun?
»Was tun Sie, Frau?« »Was geht mich das an?« »Mir bleibt nichts erspart, ich kann jetzt nur lachen!
»Wie steht das Geld?«
»Was sagen Sie, Frau, Sie haben es wohl nicht gelesen?« »Woher »?«« »Was verstehen Sie von diesen Büchern?« »Warum »Warum nicht gelesen?
Warum nicht gelesen, und nicht gelesen?«.
»Was geht mich denn das an?
Lesen Sie sich das alles!
Kaufen Sie sich über die Bücher?
»Ich lese alles was sie mir so sagen, aber ich bin ja eine ganz vernünftige Frau!« »Was steht darin geschrieben?«
Ich bin gar nicht zufrieden!
»Und was steht drin?
»Wie stehen Sie da?« »Wer weiß, was Sie über die Bücher gelesen haben.« »Warum nicht die Menschen?«
Und mit »Der Mensch!
Der Mensch!« »Der Mensch?
»Den Menschen?« »Den Mensch!
Sie können lachen!
Haben Sie sich je an mich geliehen?
»Haben Sie Ihr Buch im Ausland gelesen?
»Hat er das Buch geschrieben?« »Hat er das Bücher geschrieben, Herr Professor?
»Und hat er das Buch gelesen?« Wie gesagt!
\\
 \hline
 
\end{tabularx}
Continuation of Table \ref{table:generated_ex_1_1}. Example texts generated by the different models.
\\Part 3/3
\end{table*}

\begin{figure*}[hp]
\centering
\noindent\fbox{\begin{minipage}{40em}

		Herr Scheinfromm.
                        
                  Madam Glaubeleichten, was sagen sie? \newline

                     Frau Glaubeleichten.
                  
                  Die Wiedergeburt ist das süße Quell-Wasser
des Herzens sage ich, welches aus der Sophia urständet, und das himmlische Weltwesen gebühret. \newline

                     Herr Scheinfromm,
                   
                  ( nachdenklich.)
                  
                  Das süß - - se Quell - Was - - sehr des - - Her - -
sens - - das ist ziemlich deutlich. Wel - - ches - -
aus - - der - - So - - phi - - a - - ur - - stän - - det, - -
und - - das - - himm - - li - - sche - - Welt - - wesen
ge - - bieh - - ret. Das ist sehr schön und deutlich
erklärt. Und sie Madame?\newline

                     Frau Zanckenheimin.

                  Ich sage, es ist die Erbohrenwerdung der himmlischen Wesenheit aus der Selbstheit der animalischen Seele in dem Centro des irdischen Menschen,
und windet sich einwärts wie ein Rad.\newline

                     Herr Scheinfromm.
                  
                  Die - - Er - - boh - - ren - - wer - - dung - - der - -
hemm - - li - - schen - - We - - sein - - heit - - In Wahrheit! das ist sehr schön gesagt! Und sie Madame?\newline

                     Frau Seufzerin.      
                  
                  Es ist eine himmlische Tinktur, wodurch die
neue Seele das vegetabilische Leben der vier Elementen wegwirft, und die magische Seele, als die 
                     
Gottheit in seiner Gleichheit, nach dem Modell der
Weisheit in alle Dinge einbildet.\newline

                     Herr Scheinfromm.
                  
                  Pots tausend! das ist hoch! Eine himmlische
Tinktur, wodurch die vegetabilische Seele - - -\newline

                     Frau Seufzerin.
                  
                  Nein! die neue Seele - - -\newline

                     Herr Scheinfromm.
                  
                  Schon gut! es ist einerlei. Aber die Erklärung gefällt mir sehr.

\end{minipage}
}
\caption{Excerpt from the drama ’Die Pietisterey im Fischbein Rocke’(1736) by Luise Adelgunde Victorie Gottsched, which was one of the dramas used for training.}
    \label{fig:excerpt}

\end{figure*}

\end{document}